\pdfoutput=1

\documentclass[11pt]{article}

\usepackage{acl}
\usepackage{times}
\usepackage{latexsym}
\usepackage{textgreek}
\usepackage{mathtools}
\usepackage{amsmath}
\usepackage{txfonts}
\usepackage{xspace}
\usepackage{booktabs}
\usepackage{multirow}
\usepackage{enumitem} 
\usepackage{lipsum}
\usepackage{algorithm}        
\usepackage{pifont}
\usepackage{subcaption}
\usepackage[noend]{algpseudocode}
\usepackage{pgfplots}
\usepackage{pgfplotstable}
\usepgfplotslibrary{fillbetween} 
\usepackage{tikz}
\newcommand{\macrof}{$\mathrm{m}$-$\mathrm{F_1}$\xspace}
\newcommand{\microf}{$\mathrm{\muup}$-$\mathrm{F_1}$\xspace}

\newcommand{\cls}{\texttt{\small [cls]}\xspace}

\usepackage{microtype}

\definecolor{color_conll}{rgb}{.3,.3,1}
\tikzstyle{comick_style}=[thick, solid, #1, mark size=1.2pt, mark=*, mark options={#1, solid}]
\tikzstyle{mimick_style}=[thick, #1, dashed, mark size=1.2pt, mark=*, mark options={#1, solid}]

\title{Retrieval-augmented Multi-label Text Classification}

\date{2023}

\begin{document}
\author{ Ilias Chalkidis\thanks{\hspace{0.5em}Equal contribution.} \qquad  Yova Kementchedjhieva$^\ast$ \\
Department of Computer Science, University of Copenhagen, Denmark \\
\texttt{$\{$ilias.chalkidis,yova$\}$[at]di.ku.dk}}
\maketitle

\begin{abstract}

Multi-label text classification (MLC) is a challenging task in settings of large label sets, where label support follows a Zipfian distribution.
In this paper, we address this problem through retrieval augmentation, aiming to improve the sample efficiency of classification models. Our approach closely follows the standard MLC architecture of a Transformer-based encoder paired with a set of classification heads. In our case, however, the input document representation is augmented through cross-attention to similar documents retrieved from the training set and represented in a task-specific manner. We evaluate this approach on four datasets from the legal and biomedical domains, all of which feature highly skewed label distributions. Our experiments show that retrieval augmentation substantially improves model performance on the long tail of infrequent labels especially so for lower-resource training scenarios and more challenging long-document data scenarios.


\end{abstract}

\section{Introduction}
Multi-label text classification (MLC) is a key task in domains of critical importance, such as the legal~\cite{chalkidis-etal-2019-large} and biomedical~\cite{tsatsaronis-etal-2015-bioasq,Johnson2017} domains, where documents need to be categorized according to a set of hundreds of complementary labels. 
A common characteristic across data for these tasks is the severe class imbalance, especially problematic in cases of limited training data availability. Often times, few labels are very well attested, while many others receive limited support at training time and are thus difficult to model and assign correctly.

Few-shot learning, a related problem which assumes an extremely limited support set (often five samples or less), has been addressed in the literature through various forms of learning from nearest neighbors \cite{koch2015siamese, vinyals2016matching, snell-etal-2017, rios-kavuluru-2018-shot}. These methods are developed for single-label classification, however, and cannot be easily used to assign multiple labels.  
We propose to address the problem of limited label support in MLC through another form of nearest-neighbor learning, retrieval augmentation, which can be implemented as a direct extension of the standard approach to the task, i.e. an encoder paired with a set of classification heads.

Retrieval augmentation allows models to make predictions conditioned not just on the current input but also on information retrieved from a model-external memory. In their work on image classification, \citet{Long2022} show that augmenting inputs with similar samples retrieved from the training data results in improved classification performance, particularly on infrequent classes.
The benefits of retrieval augmentation as a form of nearest neighbor learning for text classification, on the other hand, are yet to be explored. 

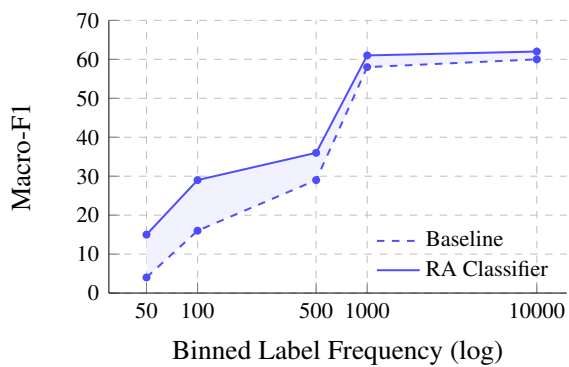
\begin{figure}[t]
\centering
\resizebox{\linewidth}{!}{
\begin{tikzpicture}

\begin{axis}[
    width=0.5\textwidth, height=.22\textheight,
    name=plot1,
        grid=major, grid style={dashed,gray!50},
    xlabel=Binned Label Frequency (log), ylabel=Macro-F1,
    xmode=log,
    ymin=0, ymax=70, xmin=30, xmax=15000,
    ytick={0, 10,20,30, 40, 50 ,60, 70},
    xtick={50, 100, 500, 1000, 10000},
    xticklabels={50, 100, 500, 1000, 10000},
    axis y line*=left, axis x line*=bottom,
    y tick label style={
        /pgf/number format/fixed,
        /pgf/number format/fixed zerofill,
        /pgf/number format/precision=0,
    },
    xticklabel style={rotate=0, font=\small},
    yticklabel style={font=\small},
    legend pos=south east,
    legend cell align=left,
    legend columns=4,
    transpose legend,
    legend style={draw=none,outer sep=0pt,inner sep=0pt,
    fill=none, font=\small, yshift=0, xshift=0pt,
        /tikz/column 2/.style={column sep=5pt}
        },
    legend entries={
        Baseline,
        RA Classifier,
        }
    ]
    \addlegendimage{no markers, color_conll, dashed, thick}
    \addlegendimage{no markers, color_conll, solid, thick}
    \addlegendimage{only marks, mark=triangle*}
    \addlegendimage{only marks, mark=square*}

    \addplot+[name path = A, mimick_style=color_conll] table[x=n_neighbors, y=macrof1, col sep=comma]{plot_data/mimic_base.csv};
    \addplot+[name path = B,comick_style=color_conll] table[x=n_neighbors, y=macrof1, col sep=comma]{plot_data/mimic_ra.csv};
    \addplot [blue!5] fill between [of = A and B];
\end{axis}

\end{tikzpicture}
}
\caption{Performance per label frequency on the MIMIC dataset \cite{Johnson2017}. Retrieval augmentation (RA) benefits lower-frequency labels more. }
\label{fig:long_tail}
\vspace{-4mm}
\end{figure}

In this work, we apply retrieval augmentation to MLC in the legal and biomedical domains and explore whether and how it improves model performance. We experiment with different design choices, evaluate the method in different data and model size settings and analyze its benefits across different label frequency bins. 
We find that retrieval augmentation benefits MLC in settings of limited data and computing availability. As shown in Figure~\ref{fig:long_tail}, this is the result of improved sample efficiency for infrequent labels.

\section{Related Work}

\paragraph{Few-Shot Learning}\hspace{-3.5mm} is a paradigm in classification tasks which targets \textit{open-class} settings, where new labels become relevant on the fly and models need to be able to assign them without re-training, based on just a handful of examples \cite{koch2015siamese, vinyals2016matching, snell-etal-2017}. The long-tail problem we address here is less constrained but related in that limited data is available for a given label. Our approach is most similar in spirit to the work of \citet{vinyals2016matching}, who compute an output distribution over labels for a given input as a weighted average of the labels of examples from a small support set. The weights are based on the similarity between the input and each of the examples in the support set, and can be computed either as simple cosine similarity or via an attention function. Their method was developed for single-label classification and cannot be trivially extended to MLC. In the space of MLC, neighbor-based learning has seen little application, limited to the use of prototype vectors for few-shot learning \cite{snell-etal-2017, rios-kavuluru-2018-shot}.

\paragraph{Retrieval Augmentation}\hspace{-3.5mm} refers to a setup where a model's prediction is condition on an input as well as supplementary input-specific information retrieved from a model-external memory. In NLP, this method has seen widespread use in knowledge-intensive tasks where the external memory is a factual source \cite{lewis2020}. When the memory is populated with training data for the task at hand, however, retrieval augmentation amounts to nearest-neighbor learning. 
While retrieval augmentation as a form of nearest-neighbor learning has not been used in text classification (to the best of our knowledge), it has shown promise for classification tasks in computer vision \citet{Long2022}.

Here, we make a logical next step in research on MLC, applying retrieval augmentation to models for the task, with the goal of improving sample efficiency and performance on infrequent labels.

\section{Methodology}
\label{sec:models}

Our retrieval-augmented MLC approach directly extends the vanilla head-based approach to the task~\cite{chalkidis-etal-2020-empirical}.

\begin{table*}[t]
    \centering
    \resizebox{\textwidth}{!}{
    \begin{tabular}{l|l|c|c|c||l}
    \toprule
        \bf Dataset & \bf Domain & \bf \# Docs (K) & \bf \# Labs & \bf \# L/D & \bf Model  \\ 
        \midrule
         BIOASQ \cite{tsatsaronis-etal-2015-bioasq} & Medical  & 80/10/10  & 112 & 9 & LexLM \cite{lexlms}\\ 
         EURLEX \cite{chalkidis2021-multieurlex} & Legal & 55/5/5 & 100 & 5 & PubMed-
BERT \cite{tinn2021pubmedbert}\\

         MIMIC \cite{Johnson2017} & Medical & 30/2.5/2.5  & 178 & 10 & PubMed-
BERT \cite{tinn2021pubmedbert}\\
        ECtHR \cite{chalkidis-et-al-2021-ecthr} & Legal & 9/1/1  & 10 & 1 & LexLM \cite{lexlms} \\
         \bottomrule
    \end{tabular}
    }
    \vspace{-2mm}
    \caption{Main characteristics of the examined datasets. We report the application domain, the number of documents across training/validation/test splits, the size of the label set, and the average number of labels per document.}
    \label{tab:datasets}
\end{table*}

\subsection{Vanilla Classifier}\label{subsec:standard_classifier}
The standard approach to MLC relies on a document encoder, paired with a set of classification heads \cite{chalkidis-etal-2020-empirical}. The document encoder, $E$, takes in a sequence of $N$ tokens, $[x_1, x_2, \dots, x_n]$, and produces a document representation, $d_i = E(x_1, x_2, \dots, x_n)$, $d\in{\rm I\!R}^{dim}$. This representation is passed to a set of $L$ classification heads, where $L$ is the size of the label set, and each classification head comprises a linear layer, $o_l\in{\rm I\!R}^{dim\times1}$, followed by a sigmoid function.

\begin{figure}[t]
    \centering
    \resizebox{.9\linewidth}{!}{
    \includegraphics{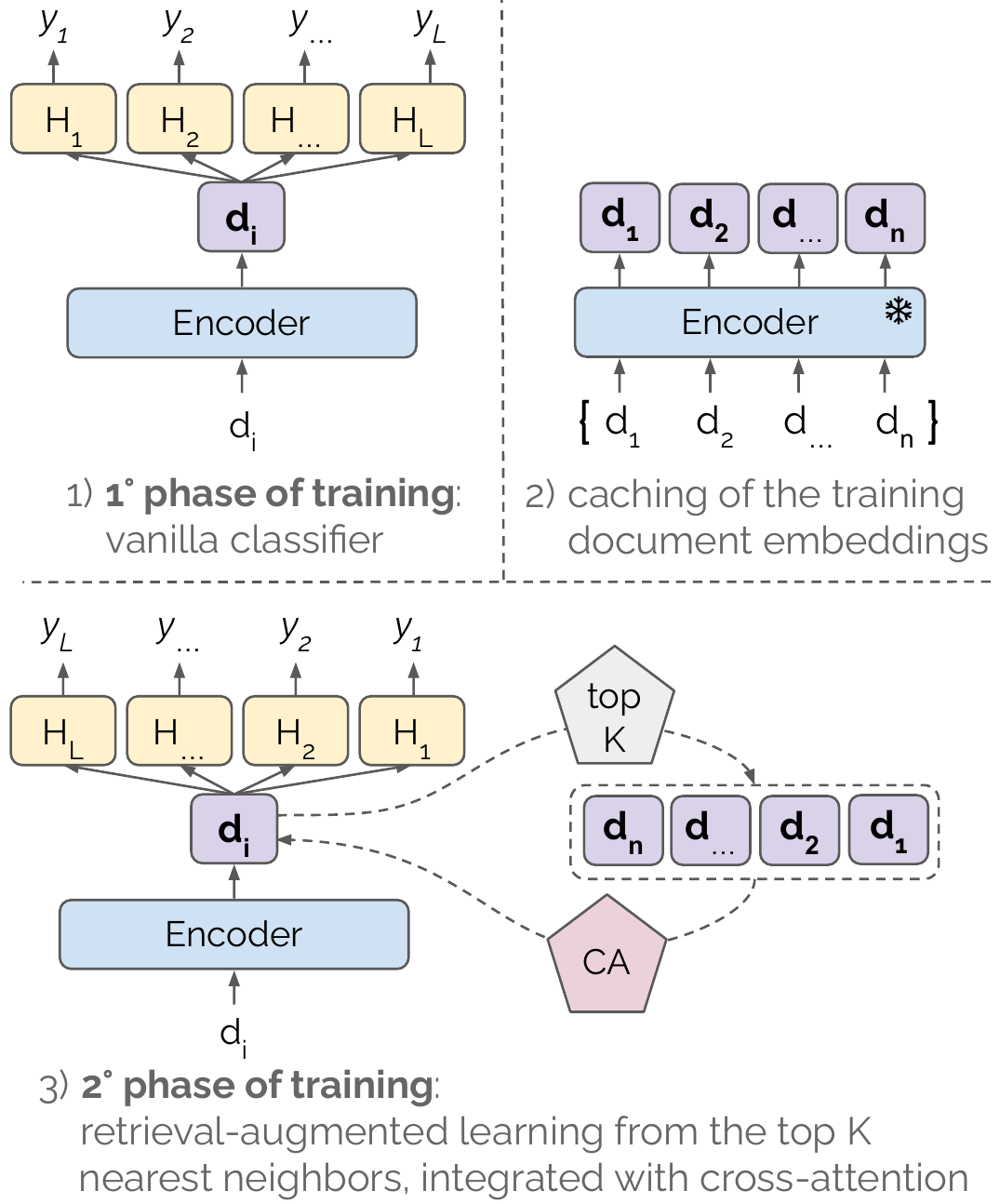}
    }
    \caption{Step-wise depiction of our approach to retrieval-augmented multi-label classification.}
    \label{fig:ra_classifier}
\end{figure}

\subsection{Retrieval-augmented Classifier} 
We extend this standard setup to allow the model to condition its predictions on the current input document as well as other similar documents retrieved from the training set. Our approach builds on the vanilla classifier both in terms of architecture and also in practical terms, since it starts from a trained vanilla classifier (see  Figure~\ref{fig:ra_classifier}). 

\paragraph{Model Architecture} Similarly to the vanilla classifier, our retrieval augmented classifier has an encoder $E$, and a set of $L$ classification heads. Additionally, it has a cross-attention (CA) module \cite{Vaswani2017}, which integrates retrieved documents into the representation of the input document (with layer normalization, LN, on top):
\begin{equation}
    \begin{aligned}
    d_i = \text{LN}(E(x_1, x_2 ..., x_n) + \text{CA}(d_1, d_2, ..., d_k))
    \end{aligned}
\end{equation}

\paragraph{Building the Retrieval Repository}
When retrieving neighbors for an input document, we want to do so based on document representations relevant to the classification task at hand. 
A vanilla classifier by design will represent documents in a task-specific manner, i.e. with relevance to the task labels. So we train a vanilla classifier in a first (preliminary) phase of training, and use its encoder to obtain representations for all training documents, caching them in a static retrieval repository.\footnote{See Appendix~\ref{sec:retrieval} for alternative strategies.} 

\paragraph{Model training}
In a second phase, we train a retrieval-augmented model, initializing its encoder and classification heads from the parameters of the phase one vanilla classifier.
Retrieval is based on cosine similarity between documents in the repository and the input document representation from the classifier's encoder (which at the start of this phase is identical to the encoder used to build the retrieval repository). Although the retrieval repository is static, the retrieval process itself is dynamic, meaning that in the course of training, the retrieval gets finetuned as well, with indirect supervision from the classification loss. 

The $K$ retrieved documents, $\{d_1, ..., d_k\}$ are integrated through cross-attention as described above.

\section{Experiments}

\subsection{Experimental Setup}

The datasets and models we use are presented in Table~\ref{tab:datasets}. For MIMIC and ECtHR, which contain long documents (1-4K tokens), we convert the vanilla Transformer-based language models into Longformer models~\cite{beltagy2020longformer}, and encode up to 2,048 tokens. For both phase one and phase two model training, we use a learning rate of 3e-5, a batch size of 32, and train models for a maximum of 100 epochs with early stopping after 5 epochs of no improvement in the macro-F1 score on the development set. Based on hyperparameter tuning conducted on a 10K sample of the BIOASQ and EURLEX datasets (see Appendix~\ref{sec:hypes} for more details), we use cross-attention of two layers, with two heads each, and set $K=4$. 

For evaluation, we are primarily interested in the macro-F1 (\macrof) score, which weighs performance on low-frequency labels equally to high-frequency ones, thus being sensitive to changes in performance on the long tail of label frequency. We additionally report micro-F1 (\microf) scores, to ensure the overall quality of our models. 

\begin{table}[t]
    \centering
    \resizebox{.9\columnwidth}{!}{
    \begin{tabular}{l|cc|cc}
    \toprule
         \bf Document & \multicolumn{2}{c|}{\bf BIOASQ} &  \multicolumn{2}{c}{\bf EURLEX} \\
         \bf Representation & \microf & \macrof  & \microf & \macrof \\
         \midrule
         Baseline  & 71.7 & 57.8  & 73.8 & 46.3\\
         \midrule
         Text  & \textbf{71.9} & \textbf{59.3} & 74.0 & \textbf{51.0} \\ 
         Labels  & 71.7  & 58.7 & \textbf{74.2} & 50.9\\ 
         Text + Labels & 71.4 & 58.8 & 74.0 & 50.8\\ 
        \bottomrule
    \end{tabular}
    }
    \caption{Development results for alternative retrieved documents representations on BIOASQ and EURLEX. }
    \label{tab:doc_repr}
\end{table}

\begin{table*}[t]
    \centering
    \resizebox{0.9\textwidth}{!}{
    \begin{tabular}{ll|cc|cc|cc|cc}
    \toprule         
         \multirow{2}{*}{\bf Model size}&\multirow{2}{*}{\bf Method} & \multicolumn{2}{c|}{\bf BIOASQ} & \multicolumn{2}{c|}{\bf EURLEX} & \multicolumn{2}{c|}{\bf MIMIC$\ast$} & \multicolumn{2}{c}{\bf ECtHR$\ast$} \\
         & & {\color{gray}  \microf } & \macrof & {\color{gray} \microf } & \macrof & {\color{gray} \microf } & \macrof & {\color{gray} \microf } & \macrof \\
         \midrule
 Base (340M) & T5Enc & {\color{gray} 75.1} & 66.0 & {\color{gray} 72.0} & 53.2 & {\color{gray} 60.5} & 31.1 &  
 {\color{gray}62.9} & 55.7\\
\midrule
\multirow{2}{*}{Base (110M)} & Baseline Classifier  &{\color{gray} 74.9} &65.7&{\color{gray} 70.7}&52.3& {\color{gray} 68.2}&35.0&{\color{gray} 70.0}&61.0\\
 & RA Classifier   & {\color{gray} 74.9} & \textbf{66.3} & {\color{gray} 70.4} & \textbf{52.9} & {\color{gray} 68.8}&\textbf{41.4}& {\color{gray} 69.3} & \textbf{64.8}\\
\midrule
\multirow{2}{*}{Large (340M)} & Baseline Classifier & {\color{gray} 76.1} & 67.5 & {\color{gray} 71.6} & \textbf{\underline{55.2}} & {\color{gray} 69.5 } & 39.8&  {\color{gray} 71.7} & 62.2\\

 & RA Classifier & {\color{gray} 76.0}&\textbf{\underline{68.0}} & {\color{gray} 70.5} & 54.2 & {\color{gray} 70.2} & \textbf{\underline{45.9}} & {\color{gray} 71.8} & \underline{\textbf{67.0}}  \\
\bottomrule
    \end{tabular}
    }
    \caption{Test results: \textbf{best score per model size}; \underline{best score overall}. \macrof scores, which weigh all labels equally, show that RA benefits performance in all cases but one. \microf scores are included for completeness. We include SOTA results from \citet{kementchedjhieva2023exploration} with T5Enc \cite{liu-etal-2022-t5enc}. $\ast$ In MIMIC and ECtHR, they used truncated versions of the documents up to 512 tokens, while we use up to 2048.}
    \label{tab:test}
    \vspace{-4mm}
\end{table*}

\subsection{Retrieved Documents Representation}
\label{sec:rdr}
A key hyperparameter we consider is the exact representation of the retrieved neighbors that would provide the best augmentation signal. While representations based on the documents' contents are optimal for retrieval, they need not be optimal for integration, since they only implicitly contain information about the documents' labels. 
Here, we compare integrating retrieved documents in terms of their document representation, a multi-hot representation of their labels, and a combination of the two (a concatenation of the two vectors, linearly projected to match the dimensionality of the main model encoder). We show results for the BIOASQ and EURLEX datasets in Table~\ref{tab:doc_repr}.

Firstly, we note that all three options result in improved performance compared to a baseline trained without retrieval augmentation---a first bit of evidence for the benefit of retrieval augmentation for MLC. 
The three variants yield highly comparable scores, but augmentation with the document representations performs best on average. This is likely because (a) the document representations by design contain a lot of information about the relevant labels, and (b) the representations of the retrieved documents occupy (approximately) the same latent space as the representation of the input document, and can thus be integrated more easily. To test point (a) above, we compute the label overlap between the input document and the $K$ retrieved neighbors (see \S~\ref{sec:lo} for details) and find it to be high indeed: .85 for BIOASQ and .92 for EURLEX.

\subsection{Main Results \& Discussion}

\paragraph{General Trends} In Table~\ref{tab:test}, we report test results for vanilla and retrieval-augmented base-size and large classifiers, trained on the full training sets of the four datasets. We find that retrieval augmentation has a mixed but overall limited impact on micro-F1 scores. Macro-F1 scores, on the other hand, show a strong trend of gains from retrieval augmentation, especially for datasets with longer documents (MIMIC and ECtHR), where boosts are observed of up to 6.4 and 5.8 points, respectively. 

\paragraph{Model Size} One may expect to find the gains of retrieval augmentation diminishing with increased model size, since a more flexible model should learn better from its training data and thus gain less from accessing that same data through retrieval. Yet in our experiments, the gains from retrieval augmentation do not seem to depend on model size in any consistent way across datasets.

\paragraph{Long Tail Labels}
For a finer breakdown of model performance on rare labels, we bin labels by their frequency in the training set, and measure macro-F1 scores within each bin. Figure~\ref{fig:long_tail} shows results for a base-size model on the MIMIC dataset. We see a clear trend of higher gains from retrieval augmentation for lower-frequency bins.

\begin{figure}[]
\centering
\resizebox{\linewidth}{!}{
\begin{tikzpicture}
\begin{axis}[
    width=0.5\textwidth, height=.22\textheight,
    name=int_cosine,
    grid=major, grid style={dashed,gray!50},
    xlabel=Number of Training Samples (K), ylabel=Macro-F1,
    ymin=25, ymax=65, xmin=0, xmax=61,
    ytick={20,30, 40, 50 ,60},
    xtick={5,10,20,55},
    axis y line*=left, axis x line*=bottom,
    y tick label style={
        /pgf/number format/fixed,
        /pgf/number format/fixed zerofill,
        /pgf/number format/precision=0,
    },
    xticklabel style={rotate=0, font=\small},
    yticklabel style={font=\small},
    legend pos=south east,
    legend cell align=left,
    legend columns=4,
    transpose legend,
    legend style={draw=none,outer sep=0pt,inner sep=0pt,
    fill=none, font=\small, yshift=0, xshift=0pt,
        /tikz/column 2/.style={column sep=5pt}
        },
    legend entries={
        Baseline,
        RA Classifier,
        }
    ]
    \addlegendimage{no markers, color_conll, dashed, thick}
    \addlegendimage{no markers, color_conll, solid, thick}
    \addlegendimage{only marks, mark=triangle*}
    \addlegendimage{only marks, mark=square*}

    \addplot+[name path = A, mimick_style=color_conll] table[x=n_samples, y=macrof1, col sep=comma]{plot_data/baseline.csv};
    \addplot+[name path = B,comick_style=color_conll] table[x=n_samples, y=macrof1, col sep=comma]{plot_data/ra_classifier.csv};
    \addplot [blue!5] fill between [of = A and B];
\end{axis}
\end{tikzpicture}
}
\caption{Models performance (\macrof) on EURLEX with respect to training data availability. }
\label{fig:more_data}
\vspace{-4mm}
\end{figure}
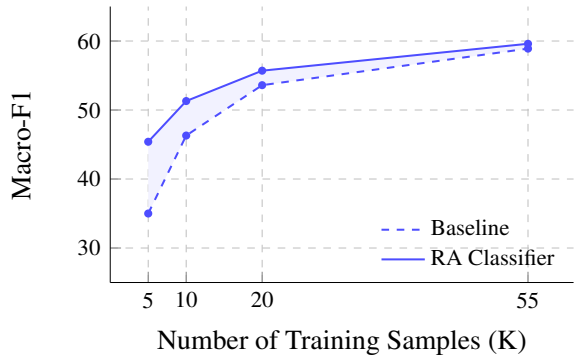

\paragraph{Training Data Availability}
\label{sec:more_data}
The datasets included in this study, for the most part, contain large training sets (see Table~\ref{tab:datasets}), which need not always be the case in real-world scenarios. Low data availability exacerbates the issue of label skewness in classification tasks, so we expect to see higher gains from retrieval augmentation in lower-resource settings. To test this, we train models on samples of the EURLEX dataset of size 5K, 10K, 20K and compare those to the models trained on the full 55K training set. Results, shown in Figure~\ref{fig:more_data}, support the hypothesis, with gains from retrieval augmentation starting off high at 5K samples and steadily diminishing as the amount of training data grows. 

\section{Conclusion}
In this work, we find that retrieval augmentation as a form of nearest neighbor learning boosts performance on mutli-label classification tasks in the legal and biomedical domains. This is the result of improved sample efficiency for infrequent labels, particularly effective in settings of limited data, and for tasks concerning long documents.

\section*{Limitations}
\paragraph{Comparison to SOTA}
In Table~\ref{tab:test} we include results from \citet{kementchedjhieva2023exploration}, who compare various approaches to MLC on the BIOASQ, EURLEX and MIMIC datasets, and find that an encoder-decoder based approach dubbed T5Enc \cite{liu-etal-2022-t5enc} is superior to the vanilla encoder-only approach, when both are initialized from a generic pre-trained language model. Here, we chose to use domain-specific language models, as a better starting point for training domain-specific classifiers. Since domain-specific encoder-decoder pre-trained language models are not available for the legal and biomedical domains, we stick to the vanilla encoder-based approach. 

\paragraph{Computational Overhead}
The method we propose carries some computational overhead compared to using a vanilla classifier. 

At training time, in addition to the vanilla classifier that we train in phase one, we also have to train a retrieval-augmented classifier in a second phase. We observe that the number of epochs until convergence in phase two training is always smaller or equal to phase one training, so we estimate the computational overhead for training at double the amount needed to just train a vanilla classifier.

In terms of the overhead from retrieval and integration through cross-attention, relevant both at training and at inference time, we find that to be negligible due to the fact that (a) we use the fast and efficient FAISS library \cite{johnson2019billion} to implement retrieval over a relatively small repository (80K documents at most) and (b) the optimal configuration for the cross-attention is relatively small (2 layers with 2 heads each) and so is the number of neighbors we integrate ($K=4$).


\bibliography{anthology,emnlp2023}

\begin{thebibliography}{24}
\expandafter\ifx\csname natexlab\endcsname\relax\def\natexlab#1{#1}\fi

\bibitem[{Beltagy et~al.(2020)Beltagy, Peters, and
  Cohan}]{beltagy2020longformer}
Iz~Beltagy, Matthew~E. Peters, and Arman Cohan. 2020.
\newblock \href {http://arxiv.org/abs/2004.05150} {Longformer: The
  long-document transformer}.
\newblock \emph{CoRR}, abs/2004.05150.

\bibitem[{Chalkidis et~al.(2019{\natexlab{a}})Chalkidis, Androutsopoulos, and
  Aletras}]{chalkidis-etal-2019-neural}
Ilias Chalkidis, Ion Androutsopoulos, and Nikolaos Aletras. 2019{\natexlab{a}}.
\newblock \href {https://doi.org/10.18653/v1/P19-1424} {Neural legal judgment
  prediction in {E}nglish}.
\newblock In \emph{Proceedings of the 57th Annual Meeting of the Association
  for Computational Linguistics}, pages 4317--4323, Florence, Italy.
  Association for Computational Linguistics.

\bibitem[{Chalkidis et~al.(2019{\natexlab{b}})Chalkidis, Fergadiotis,
  Malakasiotis, and Androutsopoulos}]{chalkidis-etal-2019-large}
Ilias Chalkidis, Emmanouil Fergadiotis, Prodromos Malakasiotis, and Ion
  Androutsopoulos. 2019{\natexlab{b}}.
\newblock \href {https://doi.org/10.18653/v1/P19-1636} {Large-scale multi-label
  text classification on {EU} legislation}.
\newblock In \emph{Proceedings of the 57th Annual Meeting of the Association
  for Computational Linguistics}, pages 6314--6322, Florence, Italy.
  Association for Computational Linguistics.

\bibitem[{Chalkidis et~al.(2021{\natexlab{a}})Chalkidis, Fergadiotis, and
  Androutsopoulos}]{chalkidis2021-multieurlex}
Ilias Chalkidis, Manos Fergadiotis, and Ion Androutsopoulos.
  2021{\natexlab{a}}.
\newblock \href {https://arxiv.org/abs/2109.00904} {{MultiEURLEX} - a
  multi-lingual and multi-label legal document classification dataset for
  zero-shot cross-lingual transfer}.
\newblock In \emph{Proceedings of the 2021 Conference on Empirical Methods in
  Natural Language Processing (EMNLP)}, Online.

\bibitem[{Chalkidis et~al.(2020)Chalkidis, Fergadiotis, Kotitsas, Malakasiotis,
  Aletras, and Androutsopoulos}]{chalkidis-etal-2020-empirical}
Ilias Chalkidis, Manos Fergadiotis, Sotiris Kotitsas, Prodromos Malakasiotis,
  Nikolaos Aletras, and Ion Androutsopoulos. 2020.
\newblock \href {https://doi.org/10.18653/v1/2020.emnlp-main.607} {An empirical
  study on large-scale multi-label text classification including few and
  zero-shot labels}.
\newblock In \emph{Proceedings of the 2020 Conference on Empirical Methods in
  Natural Language Processing (EMNLP)}, pages 7503--7515, Online. Association
  for Computational Linguistics.

\bibitem[{Chalkidis et~al.(2021{\natexlab{b}})Chalkidis, Fergadiotis,
  Tsarapatsanis, Aletras, Androutsopoulos, and
  Malakasiotis}]{chalkidis-et-al-2021-ecthr}
Ilias Chalkidis, Manos Fergadiotis, Dimitrios Tsarapatsanis, Nikolaos Aletras,
  Ion Androutsopoulos, and Prodromos Malakasiotis. 2021{\natexlab{b}}.
\newblock \href {https://aclanthology.org/2021.naacl-main.22/} {Paragraph-level
  rationale extraction through regularization: A case study on european court
  of human rights cases}.
\newblock In \emph{Proceedings of the Annual Conference of the North American
  Chapter of the Association for Computational Linguistics}, online.

\bibitem[{Chalkidis* et~al.(2023)Chalkidis*, Garneau*, Goanta, Katz, and
  Søgaard}]{lexlms}
Ilias Chalkidis*, Nicolas Garneau*, Catalina Goanta, Daniel~Martin Katz, and
  Anders Søgaard. 2023.
\newblock \href {http://arxiv.org/abs/2305.07507} {{LeXFiles and LegalLAMA:
  Facilitating English Multinational Legal Language Model Development}}.

\bibitem[{Chalkidis and S{\o}gaard(2022)}]{chalkidis-sogaard-2022-improved}
Ilias Chalkidis and Anders S{\o}gaard. 2022.
\newblock \href {https://doi.org/10.18653/v1/2022.findings-acl.192} {Improved
  multi-label classification under temporal concept drift: Rethinking
  group-robust algorithms in a label-wise setting}.
\newblock In \emph{Findings of the Association for Computational Linguistics:
  ACL 2022}, pages 2441--2454, Dublin, Ireland. Association for Computational
  Linguistics.

\bibitem[{Gao et~al.(2021)Gao, Yao, and Chen}]{gao-etal-2021-simcse}
Tianyu Gao, Xingcheng Yao, and Danqi Chen. 2021.
\newblock \href {https://doi.org/10.18653/v1/2021.emnlp-main.552} {{S}im{CSE}:
  Simple contrastive learning of sentence embeddings}.
\newblock In \emph{Proceedings of the 2021 Conference on Empirical Methods in
  Natural Language Processing}, pages 6894--6910, Online and Punta Cana,
  Dominican Republic. Association for Computational Linguistics.

\bibitem[{Johnson et~al.(2017)Johnson, Stone, Celi, and Pollard}]{Johnson2017}
Alistair~EW Johnson, David~J. Stone, Leo~A. Celi, and Tom~J. Pollard. 2017.
\newblock \href {https://www.nature.com/articles/sdata201635} {{MIMIC-III, a
  freely accessible critical care database}}.
\newblock \emph{Nature}.

\bibitem[{Johnson et~al.(2019)Johnson, Douze, and
  J{\'e}gou}]{johnson2019billion}
Jeff Johnson, Matthijs Douze, and Herv{\'e} J{\'e}gou. 2019.
\newblock Billion-scale similarity search with {GPUs}.
\newblock \emph{IEEE Transactions on Big Data}, 7(3):535--547.

\bibitem[{Koch et~al.(2015)Koch, Zemel, Salakhutdinov et~al.}]{koch2015siamese}
Gregory Koch, Richard Zemel, Ruslan Salakhutdinov, et~al. 2015.
\newblock Siamese neural networks for one-shot image recognition.
\newblock In \emph{ICML deep learning workshop}, volume~2. Lille.

\bibitem[{Lewis et~al.(2020)Lewis, Perez, Piktus, Petroni, Karpukhin, Goyal,
  K\"{u}ttler, Lewis, Yih, Rockt\"{a}schel, Riedel, and Kiela}]{lewis2020}
Patrick Lewis, Ethan Perez, Aleksandra Piktus, Fabio Petroni, Vladimir
  Karpukhin, Naman Goyal, Heinrich K\"{u}ttler, Mike Lewis, Wen-tau Yih, Tim
  Rockt\"{a}schel, Sebastian Riedel, and Douwe Kiela. 2020.
\newblock Retrieval-augmented generation for knowledge-intensive nlp tasks.
\newblock In \emph{Proceedings of the 34th International Conference on Neural
  Information Processing Systems}, NIPS'20, Red Hook, NY, USA. Curran
  Associates Inc.

\bibitem[{Liu et~al.(2021)Liu, Shakeri, Yu, and Li}]{liu-etal-2022-t5enc}
Frederick Liu, Siamak Shakeri, Hongkun Yu, and Jing Li. 2021.
\newblock \href {http://arxiv.org/abs/2110.08426} {Enct5: Fine-tuning {T5}
  encoder for non-autoregressive tasks}.
\newblock \emph{CoRR}, abs/2110.08426.

\bibitem[{Long et~al.(2022)Long, Yin, Ajanthan, Nguyen, Purkait, Garg, Shen,
  and van~den Hengel}]{Long2022}
Alexander Long, Wei Yin, Thalaiyasingam Ajanthan, Vu~Nguyen, Pulak Purkait,
  Ravi Garg, Chunhua Shen, and Anton van~den Hengel. 2022.
\newblock \href
  {https://www.amazon.science/publications/retrieval-augmented-classification-for-long-tail-visual-recognition}
  {Retrieval augmented classification for long-tail visual recognition}.
\newblock In \emph{CVPR 2022}.

\bibitem[{Nentidis et~al.(2021)Nentidis, Katsimpras, Vandorou, Krithara, Gasco,
  Krallinger, and Paliouras}]{bioasq2021}
Anastasios Nentidis, Georgios Katsimpras, Eirini Vandorou, Anastasia Krithara,
  Luis Gasco, Martin Krallinger, and Georgios Paliouras. 2021.
\newblock \href
  {https://link.springer.com/chapter/10.1007\%2F978-3-030-85251-1_18} {Overview
  of bioasq 2021: The ninth bioasq challenge on large-scale biomedical semantic
  indexing and question answering}.
\newblock In \emph{International Conference of the Cross-Language Evaluation
  Forum for European Languages (CLEF2021)}. Springer, Springer.

\bibitem[{Rios and Kavuluru(2018)}]{rios-kavuluru-2018-shot}
Anthony Rios and Ramakanth Kavuluru. 2018.
\newblock \href {https://doi.org/10.18653/v1/D18-1352} {Few-shot and zero-shot
  multi-label learning for structured label spaces}.
\newblock In \emph{Proceedings of the 2018 Conference on Empirical Methods in
  Natural Language Processing}, pages 3132--3142, Brussels, Belgium.
  Association for Computational Linguistics.

\bibitem[{Snell et~al.(2017)Snell, Swersky, and Zemel}]{snell-etal-2017}
Jake Snell, Kevin Swersky, and Richard~S. Zemel. 2017.
\newblock \href {http://arxiv.org/abs/1703.05175} {Prototypical networks for
  few-shot learning}.
\newblock \emph{CoRR}, abs/1703.05175.

\bibitem[{Tinn et~al.(2023)Tinn, Cheng, Gu, Usuyama, Liu, Naumann, Gao, and
  Poon}]{tinn2021pubmedbert}
Robert Tinn, Hao Cheng, Yu~Gu, Naoto Usuyama, Xiaodong Liu, Tristan Naumann,
  Jianfeng Gao, and Hoifung Poon. 2023.
\newblock \href {https://doi.org/https://doi.org/10.1016/j.patter.2023.100729}
  {Fine-tuning large neural language models for biomedical natural language
  processing}.
\newblock \emph{Patterns}, 4(4):100729.

\bibitem[{Tsatsaronis et~al.(2015)Tsatsaronis, Balikas, Malakasiotis, Partalas,
  Zschunke, Alvers, Weissenborn, Krithara, Petridis, Polychronopoulos,
  Almirantis, Pavlopoulos, Baskiotis, Gallinari, Artieres, Ngonga, Heino,
  Gaussier, Barrio-Alvers, Schroeder, Androutsopoulos, and
  Paliouras}]{tsatsaronis-etal-2015-bioasq}
George Tsatsaronis, Georgios Balikas, Prodromos Malakasiotis, Ioannis Partalas,
  Matthias Zschunke, Michael~R Alvers, Dirk Weissenborn, Anastasia Krithara,
  Sergios Petridis, Dimitris Polychronopoulos, Yannis Almirantis, John
  Pavlopoulos, Nicolas Baskiotis, Patrick Gallinari, Thierry Artieres, Axel
  Ngonga, Norman Heino, Eric Gaussier, Liliana Barrio-Alvers, Michael
  Schroeder, Ion Androutsopoulos, and Georgios Paliouras. 2015.
\newblock \href {https://doi.org/10.1186/s12859-015-0564-6} {An overview of the
  bioasq large-scale biomedical semantic indexing and question answering
  competition}.
\newblock \emph{BMC Bioinformatics}, 16:138.

\bibitem[{Tunstall et~al.(2022)Tunstall, Reimers, Jo, Bates, Korat, Wasserblat,
  and Pereg}]{tunstall2022}
Lewis Tunstall, Nils Reimers, Unso Eun~Seo Jo, Luke Bates, Daniel Korat, Moshe
  Wasserblat, and Oren Pereg. 2022.
\newblock \href {https://doi.org/10.48550/ARXIV.2209.11055} {Efficient few-shot
  learning without prompts}.

\bibitem[{Vaswani et~al.(2017)Vaswani, Shazeer, Parmar, Uszkoreit, Jones,
  Gomez, Kaiser, and Polosukhin}]{Vaswani2017}
Ashish Vaswani, Noam Shazeer, Niki Parmar, Jakob Uszkoreit, Llion Jones,
  Aidan~N. Gomez, Lukasz Kaiser, and Illia Polosukhin. 2017.
\newblock \href
  {https://proceedings.neurips.cc/paper/2017/file/3f5ee243547dee91fbd053c1c4a845aa-Paper.pdf}
  {Attention is all you need}.
\newblock In \emph{Proceedings of the 31st International Conference on Neural
  Information Processing Systems}, pages 6000--6010, Long Beach, California,
  USA.

\bibitem[{Vinyals et~al.(2016)Vinyals, Blundell, Lillicrap, Wierstra
  et~al.}]{vinyals2016matching}
Oriol Vinyals, Charles Blundell, Timothy Lillicrap, Daan Wierstra, et~al. 2016.
\newblock Matching networks for one shot learning.
\newblock \emph{Advances in neural information processing systems}, 29.

\bibitem[{{Yova Kementchedjhieva* and Ilias
  Chalkidis*}(2023)}]{kementchedjhieva2023exploration}
{Yova Kementchedjhieva* and Ilias Chalkidis*}. 2023.
\newblock \href {http://arxiv.org/abs/2305.05627} {{An Exploration of
  Encoder-Decoder Approaches to Multi-Label Classification for Legal and
  Biomedical Text}}.

\end{thebibliography}
\bibliographystyle{acl_natbib}
\appendix

\section{Datasets}

\label{sec:datasets}

\paragraph{EURLEX} The MultiEURLEX dataset \cite{chalkidis2021-multieurlex} consists of 65k European Union (EU) laws published on the EUR-Lex website.\footnote{\url{http://eur-lex.europa.eu/}} 
All EU laws are annotated with multiple concepts from the European Vocabulary (EuroVoc).\footnote{\url{http://eurovoc.europa.eu/}} 
EuroVoc has been used to index documents (EU laws, case law, etc.) in systems of EU institutions. We use the 2nd level of the EuroVoc taxonomy with 127 concepts (labels).

\paragraph{BIOASQ} The BIOASQ (Task A) dataset \cite{bioasq2021} consist of biomedical article abstracts released on PubMed,\footnote{\url{https://pubmed.ncbi.nlm.nih.gov}} annotated with concepts from the Medical Subject Headings (MeSH) taxonomy.\footnote{\url{https://www.nlm.nih.gov/mesh/}} MeSH comprises approx. 29k concepts of biomedical concepts (e.g., diseases, chemicals, and drugs). It is primarily used for indexing biomedical and health-related information. We use the version of the dataset used by~\citet{chalkidis-sogaard-2022-improved,kementchedjhieva2023exploration} labeled with the 2nd level of the MeSH taxonomy with 116 categories.

\paragraph{MIMIC-III} The MIMIC-III dataset \cite{Johnson2017} consists of approx.~50k discharge summaries from US hospitals. Each summary is annotated with codes (labels) from the ICD-9 taxonomy.\footnote{\url{www.who.int/classifications/icd/en/}}. The International Classification of Diseases, Ninth Revision (ICD-9) is used to assign codes to diagnoses and procedures associated with hospital utilization in the United States and is maintained by the World Health Organization (WHO). We use the version of the dataset used by~\citet{chalkidis-sogaard-2022-improved,kementchedjhieva2023exploration} labeled with 2nd level of the ICD-9 hierarchy with 184 categories.\vspace{2mm}

\paragraph{ECtHR} The ECtHR dataset \cite{chalkidis-etal-2019-neural} consists of 11K cases from the European Court of Human Rights (ECtHR).\footnote{\url{https://echr.coe.int}} The task is legal judgment prediction based on cases, where the Court decides upon allegations that a European state has breached human rights articles of the European Convention of Human Rights (ECHR).  We use the latest version of the dataset, where the total number of articles (labels) are 10.

\section{Hyperparameter Tuning}
\label{sec:hypes}
\paragraph{Cross-Attention}

As discussed in Section~\ref{sec:models}, our retrieval-augmented classifier uses cross-attention to integrate retrieved information. We perform a grid search to find the optimal number of attention layers ($\mathrm{L}\!\in\!{1,2,4}$) and attention heads per layer ($\mathrm{H}\!\in\!{1,2,4}$), given $N\!=\!32$ retrieved documents. In Table~\ref{tab:decoder_setttings}, we present development results for the BIOASQ dataset, where we observe that comparable performance for $L=2, H=2$ and $H=4,L=4$, and we adopt the former for our main experiments, as the more efficient option. 

\begin{table}[h]
    \centering
    \resizebox{.97\columnwidth}{!}{
    \begin{tabular}{l|cc|cc|cc}
    \toprule
          &  \multicolumn{2}{c|}{H=1} &  \multicolumn{2}{c|}{H=2} &  \multicolumn{2}{c}{H=4} \\
         \midrule
         & \microf & \macrof & \microf & \macrof & \microf & \macrof \\
         \midrule
L=1 & 71.4 & 58.8 & 71.5 & 58.8 & 71.5 & 58.8\\
L=2 & 71.5 & 58.8 & \textbf{71.9} & \textbf{59.3} & 71.5 & 58.9\\ 
L=4 & 71.5 & 59.0 & 71.6 & 58.9 & \textbf{71.9} & \textbf{59.3} \\
        \bottomrule
    \end{tabular}
    }
    \caption{Development results for alternative settings based on varying number of attention heads (H) and Layers (L) with N=32 neighbors on the BIOASQ dataset.}
    \vspace{-2mm}
    \label{tab:decoder_setttings}
\end{table}

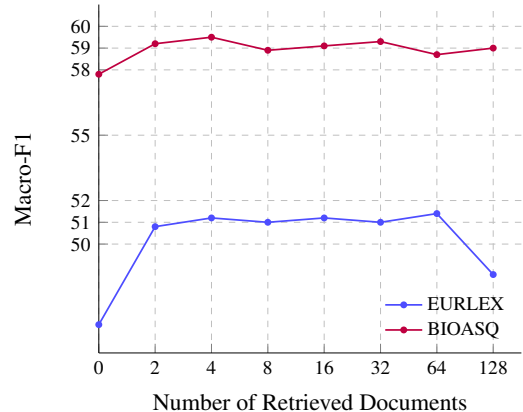
\begin{figure}[h]
\centering
\resizebox{0.9\linewidth}{!}{
\begin{tikzpicture}
\begin{axis}[
    name=int_cosine,
    grid=major, grid style={dashed,gray!50},
    xlabel=Number of Retrieved Documents, ylabel=Macro-F1,
    ymin=45, ymax=61, xmin=0, xmax=7.5,
    ytick={50, 51, 52, 55, 58, 59, 60},
    xtick={0,1,2,3,4,5,6,7},
    xticklabels={0,2,4,8,16,32,64,128},
    axis y line*=left, axis x line*=bottom,
    y tick label style={
        /pgf/number format/fixed,
        /pgf/number format/fixed zerofill,
        /pgf/number format/precision=0,
    },
    xticklabel style={rotate=0, font=\small},
    yticklabel style={font=\small},
    legend pos=south east,
    legend cell align=left,
    legend columns=1,
    transpose legend,
    legend style={draw=none,outer sep=0pt,inner sep=0pt,
    fill=none, font=\small, yshift=0, xshift=0pt,
        /tikz/column 2/.style={column sep=5pt}
        },
    legend entries={
        ,
        EURLEX,
        BIOASQ,v
        }
    ]
    \addlegendimage{no markers, color_conll, solid, thick}

    \addplot+[name path = A, comick_style=color_conll] table[x=n_neighbors, y=macrof1, col sep=comma]{plot_data/neighbors.csv};
    \addplot+[name path = B, comick_style=purple] table[x=n_neighbors, y=macrof1, col sep=comma]{plot_data/neighbors_bio.csv};
\end{axis}
\end{tikzpicture}
}
\caption{Model performance in terms of \macrof on BIOASQ and EURLEX with respect to number of retrieved documents.}
\label{fig:neighors}
\end{figure}

\paragraph{Number of Retrieved Documents}
\label{sec:neighbors}
We consider a wide range of values for the number of retrieved documents, $K$, from 2 to 128. In Figure~\ref{fig:neighors} we show results for BIOASQ and EURLEX. We note that much of the benefit of retrieval augmentation is already achieved at $K=2$. For BIOASQ, keeping $K$ low seems to work better.
For EURLEX, performance is relatively stable from $K=4$ to $64$, but drops sharply at $K=128$. This is somewhat surprising considering that even if a high value of $K$ introduces many irrelevant neighbors, the learned attention mechanism should be able to successfully ignore those.

For our main experiments, we adopt $K=4$, which yields the best performance on average between the two datasets, and is more computationally efficient compared to higher values.

\begin{table*}[t]
    \centering
    \resizebox{0.9\textwidth}{!}{
    \begin{tabular}{ll|cc|cc|cc|cc}
    \toprule         
         \multirow{2}{*}{\bf Model size}&\multirow{2}{*}{\bf Method} & \multicolumn{2}{c|}{\bf BIOASQ} & \multicolumn{2}{c|}{\bf EURLEX} & \multicolumn{2}{c|}{\bf MIMIC} & \multicolumn{2}{c}{\bf ECtHR} \\
         & & {\color{gray}  \microf } & \macrof & {\color{gray} \microf } & \macrof & {\color{gray} \microf } & \macrof & {\color{gray} \microf } & \macrof \\
         \midrule
\multirow{2}{*}{Base (110M)} & Baseline Classifier  &{\color{gray} 75.1} &66.7&{\color{gray} 76.6}&58.9& {\color{gray} 68.1}&35.4&{\color{gray} 71.9}&64.7\\
 & RA Classifier   & {\color{gray} 75.1} & \textbf{67.1} & {\color{gray} 76.4} & \textbf{59.6} & {\color{gray} 68.6}&\textbf{41.6}& {\color{gray} 70.3} & \textbf{66.0}\\
\midrule
\multirow{2}{*}{Large (340M)} & Baseline Classifier & {\color{gray} 76.3}&68.7 & {\color{gray} 76.6} & \textbf{\underline{60.0}} & {\color{gray} 69.5 } & 40.4&  {\color{gray} 73.8 } & 68.6\\

 & RA Classifier & {\color{gray}76.3 }& \textbf{\underline{69.1}} & {\color{gray} 75.9} & 59.7 & {\color{gray} 70.0} & \textbf{\underline{45.7}} & {\color{gray} 71.3} & \underline{\textbf{69.3}}  \\
\bottomrule
    \end{tabular}
    }
    \caption{Development results: \textbf{best score per model size}; \underline{best score overall}. \macrof scores, which weigh all labels equally, show that RA benefits performance in all cases but one. \microf scores are included for completeness.}
    \label{tab:test}
\end{table*}

\paragraph{Contents of the Repository} 

Given the finding that augmenting the classification model with a representation of the text of nearest neighbors works best (i.e. labels are not strictly needed, see \S~\ref{sec:rdr}), we also experiment with populating the retrieval repository with unlabeled documents, using the remainder of the full training sets of each dataset as unlabeled data.\footnote{Recall that we subsample those to 10K samples for the purposes of hyperparameter tuning.}

In Table~\ref{tab:constrained_setting}, we present results with retrieval from this larger repository versus retrieval from the labeled documents only. We find that retrieval from the labeled documents only actually works slightly better, contrary to the intuition that more data should always help. Considering that the labeled documents were also used to train the retrieval encoder, it likely is the case that their representations are more accurate than those of the unlabeled documents, i.e. that it is either difficult for the retrieval encoder to generalize to unseen data, or for the classification model to extract accurate signal from the representations of unlabeled documents. 

\begin{table}[h]
    \centering
    \resizebox{\linewidth}{!}{
    \begin{tabular}{l|cc|cc}
    \toprule
         Retrieval data &  \multicolumn{2}{c|}{\bf BIOASQ} & \multicolumn{2}{c}{\bf EURLEX} \\
         & \microf & \macrof & \microf & \macrof  \\
         \midrule
         Labeled only &  72.0 & 59.5 & 74.1 & 51.2  \\
         Labeled + Unlabeled & 72.1 & 59.8 & 73.8 & 51.0   \\
\bottomrule
    \end{tabular}
    }
    \caption{Development results for contents of the retrieval repository.}
    \label{tab:constrained_setting}
\end{table}

\section{Retrieval}

\subsection{Label Overlap}\label{sec:lo}
To shed more light on the utility of retrieved documents, we compute a Label Overlap (LO) ratio as follows:

\begin{table}[t]
    \centering
    \resizebox{\columnwidth}{!}{
    \begin{tabular}{l|ccc|ccc}
    \toprule
         \multirow{2}{*}{\bf Retriever} & \multicolumn{3}{c|}{\bf BIOASQ} &  \multicolumn{3}{c}{\bf EURLEX} \\
          & LO & \microf & \macrof & LO & \microf & \macrof \\
         \midrule
         Baseline & - & 71.7 & 57.8  & - & 73.8 & 46.3 \\
         \midrule
         CLS &  .85 &  72.0 & 59.5 & .92 &  74.2 & 51.3  \\
         SimCSE & .61 &  71.5 & 57.5 & .52 & 74.4 & 51.5 \\
         SimCSE+ & .57 & 71.3 & 57.2 & .76 & 74.4 & 51.1 \\
        \bottomrule
    \end{tabular}
    }
    \caption{Development Results for alternative retrievers on BIOASQ and EURLEX. We report the Label Overlap (LO) ratio between the target and retrieved documents, micro- (\microf), and macro-F1 (\macrof).}
    \label{tab:retriever}
\end{table}

\begin{equation}
   \mathrm{LO} = \frac{1}{N}\sum_{i=1}^{n} \frac{1}{K}\sum_{j=1}^{k} \frac{|L_i \cap L_j|}{\mathrm{min}(|L_i|, |L_j|)}
\end{equation}

\noindent where $L_i$ is the set of labels of the $i$th input document of the training set, and $L_j$ the set of labels of the $j$th retrieved document, $N$ being the number of training samples, and $K$ being the number of neighbors.
\subsection{Retrieval Encoder}
\label{sec:retrieval}
 
To determine the value of the first phase of training which serves primarily to train a document encoder, we explore two alternative approaches:
(a) Using \emph{unsupervized SimCSE}~\cite{gao-etal-2021-simcse}, a contrastive learning method to pre-train a document embedder, in which case positive pairs come from different views of the same document, while all other documents are considered for negative sampling,  and (b) Using \emph{supervized SimCSE}~\cite{gao-etal-2021-simcse,tunstall2022}, in which case similarly labeled documents are considered as positive pairs, and negative otherwise.
Similarly to before, we consider \cls pooling to represent document embeddings, and we rank documents based on their cosine similarity with the input document, as represented by the same document embedder. In Table~\ref{tab:retriever}, we report results for the alternative retriever encoders for the BIOASQ and EURLEX datasets. 

We find that these alternative encoders lead to worse classification performance compared to our proposed retriever for BIOASQ, while for EURLEX performance is comparable between the three approaches. 
In Table~\ref{tab:retriever}, we observe that the encoder trained within a \emph{vanilla classifier} results in retrieval of documents with an average label overlap ratio of .85 and .92 respectively for BIOASQ and EURLEX, while the retrievers based on SimCSE lead to a much lower label overlap. Interestingly, this value correlates with the gains from retrieval augmentation for BIOASQ but not for EURLEX. 

\end{document}